\pdfoutput=1
\documentclass[twocolumn]{article}

\usepackage{subcaption}
\usepackage{calc}
\usepackage{amssymb}
\usepackage{amstext}
\usepackage{amsmath}
\usepackage{amsthm}
\usepackage{multicol}
\usepackage{pslatex}

\usepackage{graphicx}
\usepackage{algorithm}
\usepackage{algpseudocode}
\usepackage{siunitx}

\usepackage{pdfpages}

\begin{document}

\title{Accelerated RRT* and its evaluation on Autonomous Parking}

\author{
Jiri Vlasak$^1$$^2$
, Michal Sojka$^2$
and Zden\v{e}k Hanz\'{a}lek$^2$
}

\date{{\small
$^1$Faculty of Electrical Engineering, Czech Technical University in Prague
\\
$^2$Czech Institute of Informatics, Robotics and Cybernetics, Czech Technical University in Prague
\\
%Email: jiri.vlasak.2@cvut.cz
\{jiri.vlasak.2, michal.sojka, zdenek.hanzalek\}@cvut.cz
}}

\null
\includepdf[pages=1,fitpaper,noautoscale]{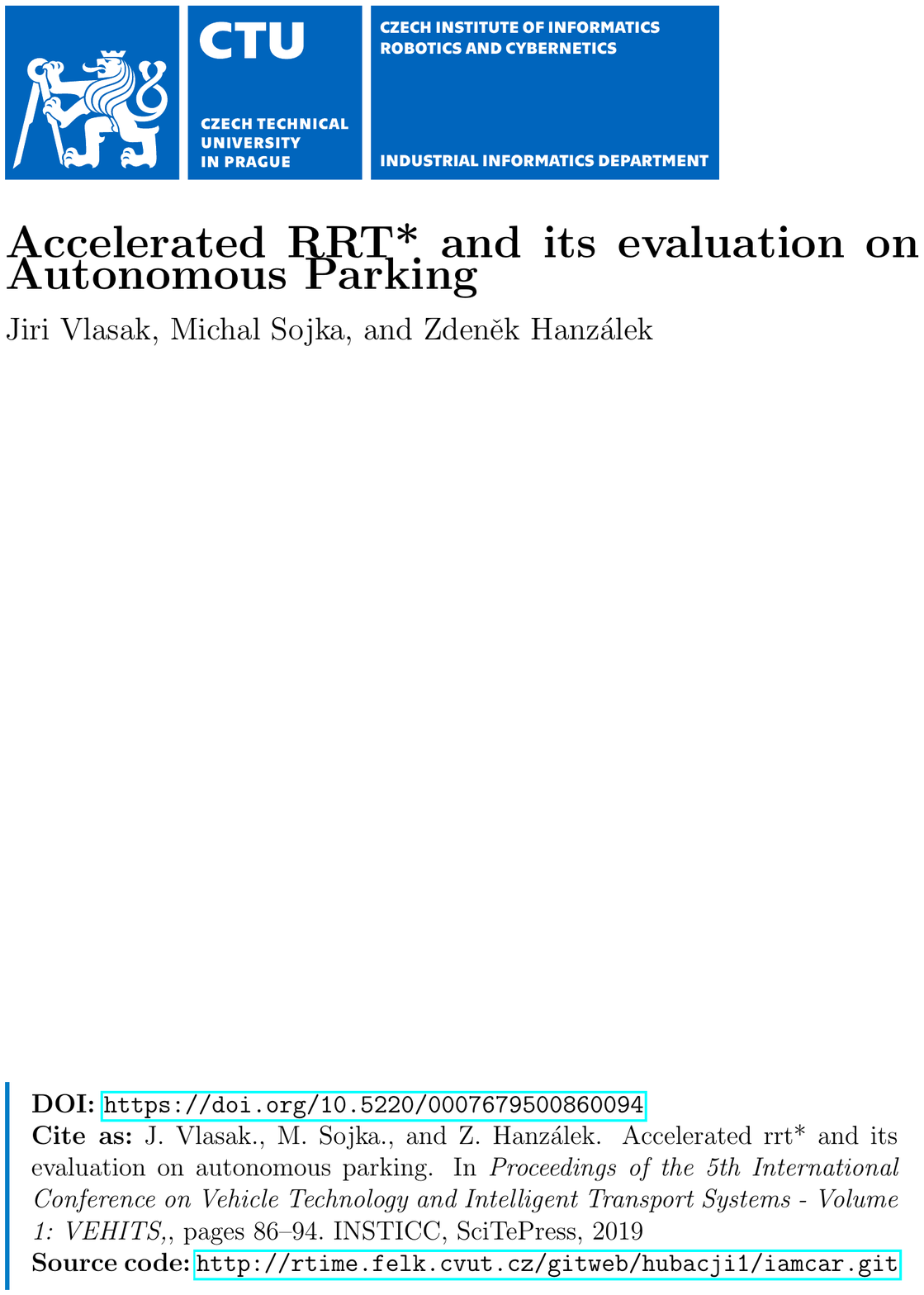}
\maketitle \normalsize \vfill

\abstract{
Finding a collision-free path for autonomous parking is usually performed by
computing geometric equations, but the geometric approach may become unusable
under challenging situations where space is highly constrained.  We propose an
algorithm based on Rapidly-Exploring Random Trees Star (RRT*), which works even
in highly constrained environments and improvements to RRT*-based algorithm
that accelerate computational time and decrease the final path cost.  Our
improved RRT* algorithm found a path for parallel parking maneuver in
\SI{95}{\%} of cases in less than \SI{0.15}{seconds}.

}

\section{\uppercase{Introduction}}

\noindent Modern cars are commonly equipped with parking assistants that can
perform parallel or perpendicular parking maneuvers. Parking is a relatively
easy task as the movement is slow and the car dynamics might be neglected.
Usually, geometric equations are used for planning these maneuvers. A geometric
approach has limitations when applied in unexpected environments or when more
than a simple parking maneuver has to be planned. In this paper we address the
cases, when more advanced planners need to be used, and one of the problems
experienced by those complex planners is their computational complexity.

For this paper, we define the parking problem as finding a collision-free path
from an initial car position (i.e., \(x\), \(y\), and \(heading\)) to the goal
position under the presence of an arbitrary number of known static obstacles.
The path may consist of an arbitrary number of path segments alternating
forward and backward drives of the car. We are interested in a close to optimal
parking maneuver path in the sense of path length respecting the kinematic
constraints of the car.

In this paper, we propose an RRT*-based algorithm to solve the autonomous
parking problem, which we define more formally in Section~\ref{sec:problem}.
Contrary to well-known A* algorithm, RRT* algorithm does not need space
discretization.  Also, it handles nonholonomic constraints by design. The RRT*
algorithm searches the state space by creating a tree structure that represents
possible paths. RRT*-based algorithms were successfully applied to a wide range
of planning problems from the robot, vehicle, and aerial domains. However, they
were also used in not such apparent problems as tunnel detection in proteins
from the field of molecular biology.

Our algorithm uses Reeds and Shepp curves for particular path segments when
building the tree and Euclidean distance as a metric for the nearest neighbor
search. We complemented the RRT* algorithm with an optimization procedure based
on the Dijkstra algorithm used to reduce the number of the path segments and to
lower the cost of the path connecting initial and goal pose.

The main contributions of this paper are:

\begin{itemize}
        \item Minimization of the path cost with an optimization procedure
        based on the shortest path by Dijkstra algorithm.
        \item Speed up of the RRT* path search with the nearest neighbor
        heuristics.
\end{itemize}

In our experiments (see Section~\ref{sec:ce}), we compare multiple cost
functions of the nearest neighbor search and show that the fastest approach to
find the path is to use the Euclidean distance as the cost function in the
nearest neighbor search (see Figure~\ref{fig:noth}). We also evaluate the
effectiveness of our optimization procedure based on the Dijkstra algorithm and
show (see Figure~\ref{fig:noch}) that it significantly improves the cost of the
path even when compared to other algorithms such as RRT*-Smart. In
Section~\ref{sec:concl} we summarize our results. The source code of our
algorithm is
available~\footnote{http://rtime.felk.cvut.cz/gitweb/hubacji1/iamcar.git}.

\subsection{Related Works}

\noindent A common approach to solve a parking problem is to split the task to
the environment detection, the path planning, and the path execution. In this
paper, we consider the path planning part.

Typical parking problems can be classified into two classes: parallel parking
and perpendicular parking. Some publications consider only parallel parking
\cite{Gupta2010}, \cite{Cheng2013}, \cite{Vorobieva2013}, or only perpendicular
parking \cite{Petrov2015}. In this paper, we propose a universal method which
considers obstacles of arbitrary shape.

Many published approaches use Reeds and Shepp curves~\cite{Reeds1990} for path
planning~\cite{Lee2006} without considering obstacles. In~\cite{Fraichard2004},
the authors present Continuous-Curvature Paths that extend the Reeds and Shepp
line segments and circular arcs with clothoid arcs. Resulting paths have
continuous curvature, so a car that follows a path does not have to stop to
change orientation of the wheels.  Continuous-Curvature Paths have been used in
\cite{Muller2007}, \cite{Vorobieva2013}, \cite{Cheng2013}, and \cite{Yi2017}.
In~\cite{Kim2010}, the authors use two basic motions to create a set of
motions. Finally, \cite{Hsu2008}, \cite{Gupta2010}, and \cite{Liang2012}
describe parking using paths generated with two circles geometry.

However, in real-life situations, a typical parking scenario may be disturbed
by sloppy parked neighbor car, temporary parked bike, non-standard parking slot
shape, or other unspecified constraints. Therefore, when a parking slot is
detected, evaluation of a situation may fail, and an approach based on
geometric equations may become unusable in such a case.

In this paper, we propose RRT*-based algorithm which can handle complex parking
situations. Rapidly-Exploring Random Trees (RRT) \cite{LaValle1998} is a
randomized algorithm that can handle nonholonomic constraints. Although RRT is
probabilistically complete (with probability \SI{1}{}, the algorithm converges
to the solution, as time tends to infinity), it is not asymptotically
optimal~\cite{Karaman2011}.  Therefore, Karaman and Frazzoli proposed the RRT*
algorithm, which converges to an optimal solution as time tends to infinity.
In~\cite{Islam2012}, the authors improved the RRT* algorithm by using path
optimization and intelligent sampling and named the resulting algorithm
RRT*-Smart. After the initial path is found, RRT*-Smart converges to the
optimum faster than RRT*. In our approach, we stop the RRT* algorithm when a
path is found, and then we optimize the path by Dijkstra algorithm.

\section{\uppercase{The parking problem}}
\label{sec:problem}

\noindent In this section, we define the problem and terminology used
throughout this paper.

\begin{figure}[!htbp]
        \vspace{-0.2cm}
        \centering{
         \includegraphics[width=\linewidth]{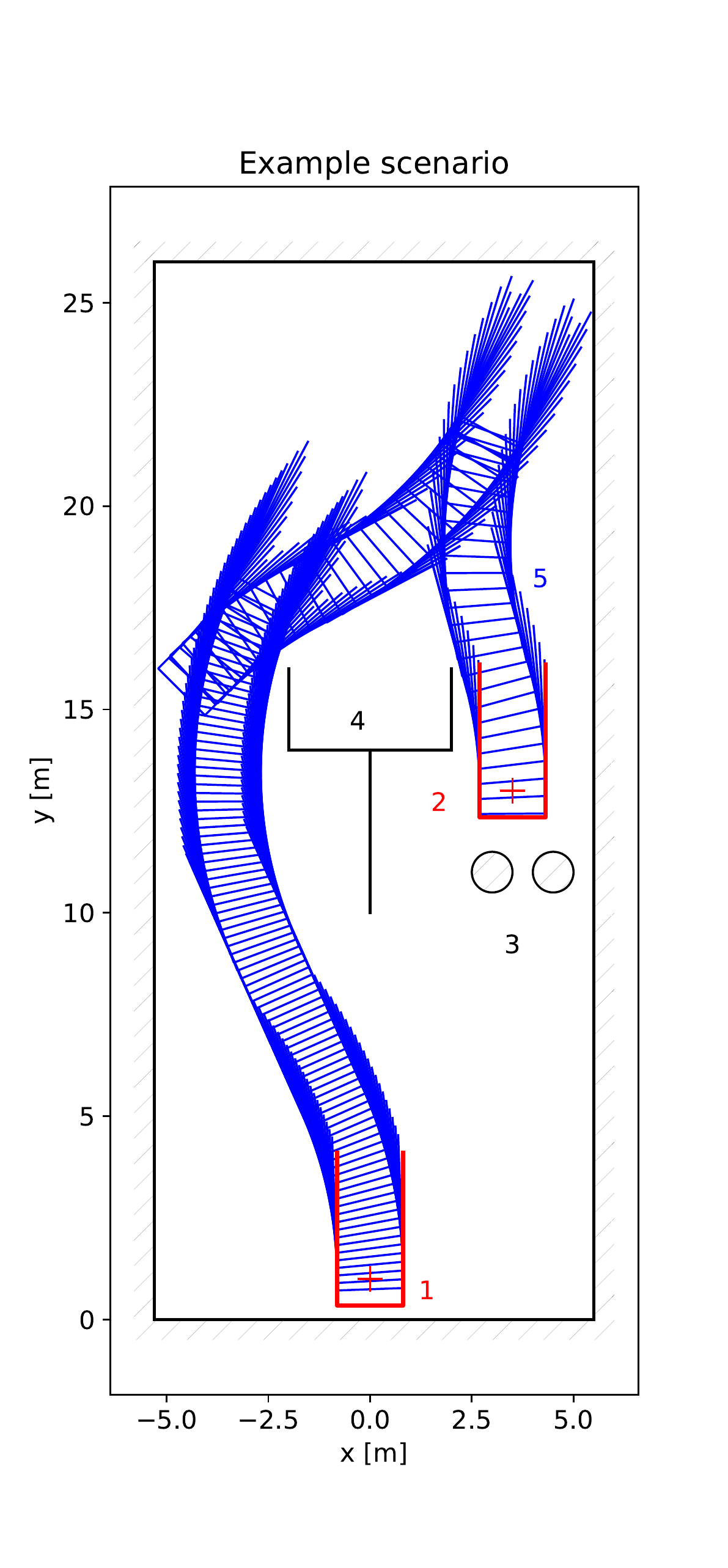}}
        \caption{Example scenario with the init pose (1), the goal pose (2),
        two circle obstacles (3), the obstacle compound of line segment
        obstacles (4), and the final path (5).}

        \label{fig:exsc_fp}
        \vspace{-0.1cm}
\end{figure}

A {\em pose} is a triplet \(p = (x, y, \theta)\), where \(x, y\) are cartesian
coordinates and \(\theta\) is a heading.

A {\em search space} is a set of poses \(S = \{(x, y, \theta) \mid x\in [XMIN,
XMAX], y\in [YMIN, YMAX], \theta\in [0, 2\pi)\}\), where \(XMIN\), \(XMAX\),
\(YMIN\), and \(YMAX\) are borders of search space.

A {\em scenario} is a quintuple \(s = (S, p_{init}, p_{goal}, O_C, O_S)\),
where \(S\) is a search space, \(p_{init}, p_{goal}\) are init and goal poses,
and \(O_C\) resp. \(O_S\) are sets of circle obstacles resp. line segment
obstacles. We can see an example scenario with the final path connecting
initial and goal pose in Figure~\ref{fig:exsc_fp}. Example scenario also
demonstrates segment obstacles (borders), circle obstacles, and the complex
obstacle of arbitrary shape (compound of line segment obstacles).

{\em Circle obstacle} is a triplet \(o_c = (x, y, r)\), where \(x, y\) are
cartesian coordinates of the center, and \(r\) is the radius. {\em Line
segment} obstacle is a quadruple \(o_s = (x_1, y_1, x_2, y_2)\), where \(x_1,
y_2\) are coordinates of the line segment start and \(x_2, y_2\) are
coordinates of the line segment end.

A {\em car} is a quadruple \(c = (l, w, R, b)\), where \(l\) is a length of the
car, \(w\) is a width of the car, \(R = \frac{1}{\kappa}\) is car minimum
turning radius, \(\kappa\) is curvature, and \(b\) is car wheelbase (the
distance between front and rear axles). In Figure~\ref{fig:exsc_fp}, red
crosses represent \(x, y\) coordinates of $init$ and $goal$ poses.  The {\em
U-Shape} frame represents length \(l\), width \(w\), and pose heading.
Finally, example obstacles are hatched.

A {\em path} from pose \(a\) to pose \(b\) is a sequence of poses \(P_{a,b} =
\{p_i \mid i\in \{0, 1, ..., n-1\}, p_0 = a, p_{n-1} = b\}\), such that \(P\)
satisfies kinematic constrains given by car \(c\).

The \(collide(p, O)\) function returns \(True\) when a car \(c\) positioned at
pose \(p\) is inside arbitrary obstacle \(o\in O\), or the frame of car \(c\)
collides with this obstacle. Otherwise, the function returns \(False\).

The \(collide(P, O)\) function returns \(True\) when for any pose \(p\in P\)
the \(collide(p, O)\) returns \(True\). Otherwise, the function returns
\(False\).

The \(cost(P)\) is a {\em path cost} defined in Equation~\ref{eq:cost}, where
\(RSDist(a, b)\) is Reeds \& Shepp distance from pose \(a\) to pose \(b\).

\begin{equation}
\label{eq:cost}
cost(P) = \sum_{i=0}^{i=n-2} RSDist(p_i, p_{i+1})
\end{equation}

We define {\em final path} \(P_F\) in Equation~\ref{eq:PF}, where \(P_{all} =
\{P_{a, b}\mid a=p_{init}, b=p_{goal}, \neg collide(P, O)\}\). An example of
the {\em final path} is in Figure~\ref{fig:exsc_fp}.

\begin{equation}
\label{eq:PF}
P_F=\underset{P\in P_{all}}{{\arg\min}_{P}}\ cost(P)
\end{equation}

\section{\uppercase{RRT*}}
\label{sec:rrt}

\noindent Rapidly-Exploring Random Tree Star (RRT*) is an asymptotically
optimal randomized algorithm to solve path planning problems, such as the
parking problem defined in Section~\ref{sec:problem}.

RRT* uses a tree data structure that represents poses and paths, it handles
nonholonomic constraints and can hold general restrictions on \(p_{init}\) and
\(p_{goal}\) poses, or obstacles. Therefore, the RRT* should be able to solve
the unpredictable, real-life scenarios. We can see basic RRT* pseudocode
(lines~\ref{alg:rrts3} to~\ref{alg:rrts4}) as part of complete RRT*-based
Algorithm~\ref{alg:rrtstar}.

The fundamental element of RRT* is a \(node\). The \(node\) is a pose extended
with \(parent\) (the pointer to the predecessor \(node\)), \(children\) (the
array of successor \(nodes\)), and cumulative cost \(ccost =
cost(P_{p_{init},node})\). As \(node\) is extension to pose, we may update our
definition of path \(P_{a,b} = \{p_i \mid i\in \{0, 1, ..., n-1\}, p_0 = a,
p_{n-1} = b\}\), such that \(a\), \(b\), and \(p_i\) are nodes, where \(p_i\)
is parent of \(p_{i+1}\). We use a \(path\) as the sequence of poses or
\(nodes\) interchangeably.

In RRT* algorithm, all \(nodes\) are stored in tree data structure
\(\mathcal{T} = (root, V, E)\), where \(root\) node corresponds to \(p_{init}\)
pose, \(V\) is set of nodes, \(E\) is set of edges, and \(\forall n_1, n_2\in
V: \{n_1, n_2\}\in E \Leftrightarrow n_1\text{ is the parent of }n_2\).

\begin{figure}[!htbp]
        \vspace{-0.2cm}
        \centering{
         \includegraphics[width=\linewidth]{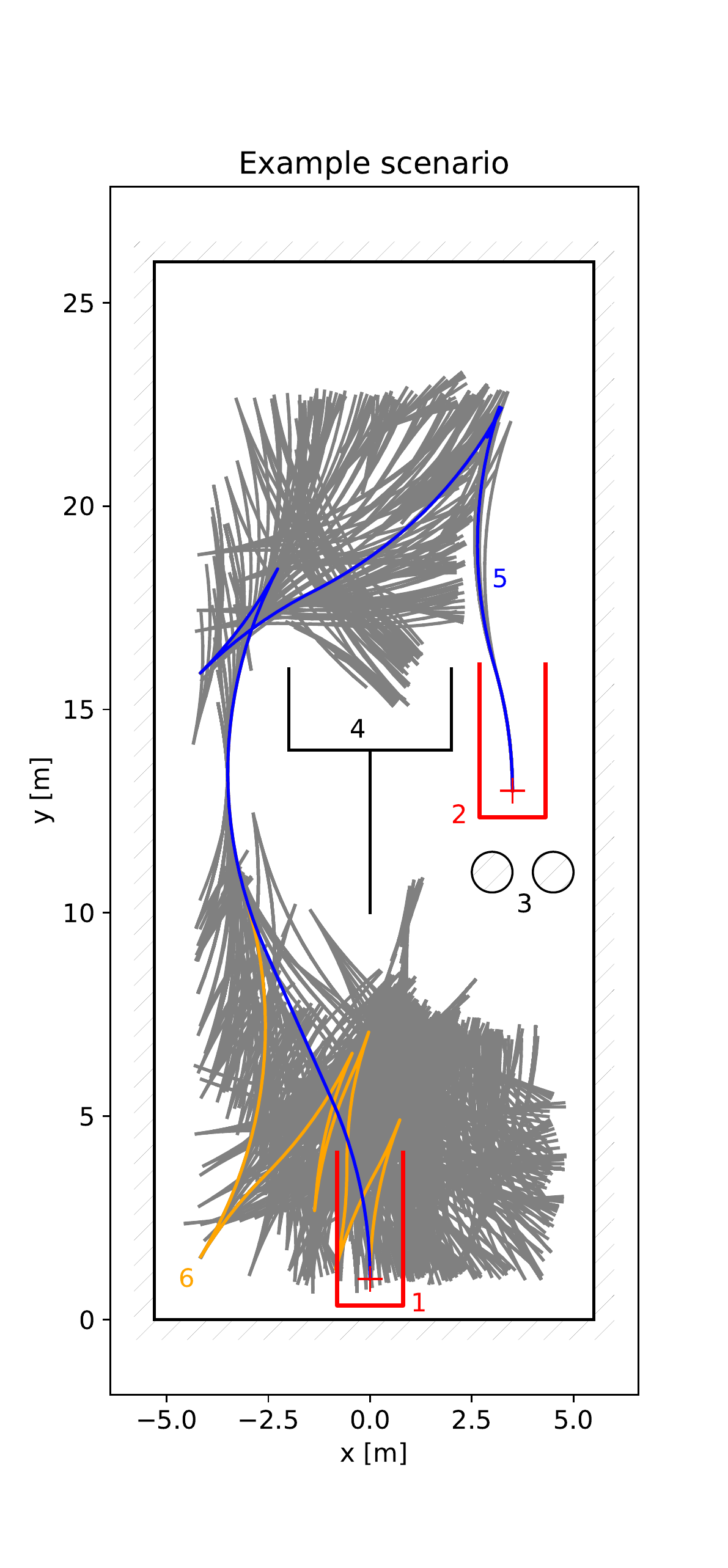}}
        \caption{Example scenario with the init pose (1), the goal pose (2), two circle
        obstacles (3), the obstacle compound of line segment obstacles (4), the final
        path (5), the final path before optimization (6), and line segments and
        circle segments (gray).}

        \label{fig:exsc_cr}
        \vspace{-0.1cm}
\end{figure}

\subsection{Basic Procedures}
\label{sec:rrt_bp}

\noindent In this section, we describe the basic procedures of RRT* used to
build \(\mathcal{T}\) data structure.

\textsc{RandomSample} procedure returns a node with a pose from search space
\(S\), where \(x\), \(y\), and \(\theta\) are randomly generated.

\textsc{Cost(\(nn, rs\))} function is a metric used in RRT*.

\textsc{NearestNeighbor(\(rs\))} procedure searches for a \(node\) with the
lowest \textsc{Cost(\(node, rs\))} in \(\mathcal{T}\).

\textsc{Steer(\(nn, rs\))} procedure returns a path \(P_{nn, rs}\). We can see
the results of \textsc{Steer} procedure in Figure~\ref{fig:exsc_cr} (gray).

\textsc{NearNodes(\(ns, dist\))} procedure returns a set of nodes \(nns\) from
\(\mathcal{T}\), such that \(\forall n\in nns:\textsc{Cost}(n, ns)<dist\).

\textsc{Connect(\(ns, nns\))} procedure searches in near nodes (\(nns\)) for
the best candidate node to expand \(\mathcal{T}\) towards the \(ns\). The best
candidate node is the node in \(\mathcal{T}\) that minimizes the cumulative
cost of \(ns\) when it becomes the parent of the \(ns\). The path from the best
candidate node to the \(ns\) must be free of collisions. If the best candidate
node is found, the \(na\) is added as a child of the best candidate node, and
\textsc{Connect(\(ns, nns\))} returns \(True\). Otherwise, the procedure
returns \(False\).

\textsc{Rewire(\(ns, nns\))} procedure checks if for any \(node\) in \(nns\)
there is a path with lower cumulative cost via \(ns\). And swaps parents if so.
This procedure along with \textsc{Connect(\(ns, nns\))} ensures the
asymptotical optimality of RRT*.

\textsc{GoalFound} returns \(True\) if \(p_{goal}\in \mathcal{T}\) and
\(False\) otherwise.

\textsc{Collides(\(na, ns\))} returns \(True\) if the path \(P_{na, ns}\)
collides with any obstacle of scenario, and \(False\) otherwise.

\begin{algorithm}[!htbp]
        \caption{Accelerated RRT*}
        \label{alg:rrtstar}
        \begin{algorithmic}[1]
        \State Input:\begin{itemize}
                \item initial pose
                \item goal pose
                \item array of obstacles
        \end{itemize}
        \State Output:\begin{itemize}
                \item \(True\) if goal pose reached, \(False\) otherwise
                \item array of paths connecting initial and goal pose
        \end{itemize}
                \Procedure{RRT*}{}
                \While{$\Call{Elapsed}{} < TMAX$}\label{alg:rrts3}
                \State $rs\gets \Call{RandomSample}{}$
                \State $nn\gets \Call{NearestNeighbor}{rs}$
                \State $pn\gets nn$
                \State $newNodes\gets \emptyset$
                \For{$ns\gets \Call{Steer}{nn, rs}$}
                        \State $nns\gets pn\cup\Call{NearNodes}{ns, dist}$
                        \If{$\Call{Connect}{ns, nns}$}
                                \State $\Call{Rewire}{ns, nns}$
                                \State $newNodes\gets newNodes \cup ns$
                                \If{$\Call{GoalFound}{}$}
                                        \State \textbf{break} while
                                \EndIf
                                \State \(pn\gets ns\)
                        \EndIf
                \EndFor\label{alg:rrts4}
                \For{$na\gets newNodes$}\label{alg:rrts5}
                        \State $pn\gets na$
                        \For{$ns\gets \Call{Steer}{na, goal}$}
                                \If{$\Call{Collide}{pn, ns}$}
                                        \State \textbf{break}
                                \EndIf
                                \State \(pn.children\gets pn.children\cup ns\)
                                \If{\Call{GoalFound}{}}
                                        \State \textbf{break} while
                                \EndIf
                                \State $pn\gets ns$
                        \EndFor
                \EndFor\label{alg:rrts6}
                \EndWhile
                \If{\Call{GoalFound}{}}
                        \State\Call{OptPath}{}\label{alg:rrts_1}
                \EndIf\label{alg:rrts_2}
                \State \textbf{return} \Call{GoalFound}{}
                \EndProcedure
        \end{algorithmic}
\end{algorithm}

\subsection{Implementation}
\label{sec:rrt_i}

\noindent Our \textsc{RandomSample} procedure samples randomly from the whole
space \(S\) (including obstacles). We use OMPL~\cite{Sucan2012} implementation
of Reeds and Shepp~\cite{Reeds1990} optimal paths for \textsc{Steer} and
\textsc{Cost} functions.  \textsc{NearNodes}, \textsc{Connect} and
\textsc{Rewire} procedures work the same as in~\cite{Karaman2011}.

For two nodes we implemented auxiliary \textsc{IsNear(\(n_1, n_2\))} function
that returns $True$ if $n_1$ is within the predefined Euclidean distance from
$n_2$ ($GFDIST$) and the difference between headings of $n_1$ and $n_2$ is less
than the specified angle ($GFANGLE$). We use this function to specify if the
goal was found, the \textsc{Steer} procedure reached $rs$, or if two nodes are
the same.  For computational experiments in Section~\ref{sec:ce}, we used
$GFDIST = 0.05$ and $GFANGLE = \frac{\pi}{32}$.

In each iteration of RRT*-based algorithm, there is an expansion of
\(\mathcal{T}\) towards the \(p_{goal}\) (see lines~\ref{alg:rrts5}
to~\ref{alg:rrts6} in Algorithm~\ref{alg:rrtstar}) as used
in~\cite{Kuwata2008}. We added path optimization procedure to RRT*-based
algorithm (see line~\ref{alg:rrts_1} in Algorithm~\ref{alg:rrtstar}) that is
run when the goal is found as explained in Section~\ref{sec:po}. We can see an
example of optimized final path (5) and final path before optimization (6) in
Figure~\ref{fig:exsc_cr}.

\section{\uppercase{RRT* Improvements}}
\label{sec:rrtsi}

\noindent In this section, we introduce our improvement to nearest neighbor
search and details about path optimization procedure.

\subsection{Nearest Neighbor}

\begin{algorithm}[!htbp]
        \caption{Nearest neighbor search}
        \label{alg:nn}
        \begin{algorithmic}[1]
                \State $IYSIZE$\Comment{nn structure size}
                \State $IYSTEP$\Comment{increment distance}
                \State $nodes[IYSIZE]$\Comment{array of lists of nodes}
                \label{alg:nn_1}
                \\
        \State Input:\begin{itemize}
                \item \(node\) to be added to data structure
        \end{itemize}
        \State Output:\begin{itemize}
                \item data structure of nodes
        \end{itemize}
                \Procedure{AddIY}{$node$}\label{alg:nn_2}
                \State $iy\gets \lfloor \frac{node.y}{IYSTEP} \rfloor$
                \State $nodes[iy]\gets nodes[iy] \cup node$
                % \State \Call{Push}{$nodes[iy], node$}
                \EndProcedure\label{alg:nn_3}
                \\
        \State Input:\begin{itemize}
                \item \(node\) to be searched
        \end{itemize}
        \State Output:\begin{itemize}
                \item the nearest neighbor of \(node\)
        \end{itemize}
                \Procedure{NearestNeighbor}{$node$}\label{alg:nn_4}
                \State $iy\gets \lfloor \frac{node.y}{IYSTEP} \rfloor$
                \State $nn\gets NULL$\Comment{nearest neighbor}
                \State $c_{min}\gets \infty$\Comment{minimum cost}
                \State $as\gets 0$\Comment{array step}
                \While{$c_{min} > as \cdot IYSTEP$}
                        \State $i\gets \max(iy-as, 0)$
                        \State $j\gets \min(iy+as, IYSIZE-1)$
                        \For{$n\in nodes[i]\cup nodes[j]$}
                                \If{$\Call{EDist}{n, node} < c_{min}$}
                                \State $c_{min}\gets \Call{EDist}{n, node}$
                                \State $nn\gets n$
                                \EndIf
                        \EndFor
                        \State $as\gets as + 1$
                \EndWhile
                \EndProcedure\label{alg:nn_5}
        \end{algorithmic}
\end{algorithm}

\noindent Because the nearest neighbor procedure returns a node with the lowest
cost, such a node is a good candidate for tree expansion. The pseudocode of the
nearest neighbor search is outlined in Algorithm~\ref{alg:nn}. To improve the
performance of finding the nearest neighbor, we use a $nodes$ data structure
(the array of linked lists of nodes) defined in line~\ref{alg:nn_1}. The
$nodes$ data structure allows us to split search space \(S\) along the
\(y\)-axis (\(y-axis\) suits better for parallel parking scenario we
experimented with in Section~\ref{sec:ce}), so we can compare nodes within
multiples of \(IYSTEP\) (increment distance based on $nodes$ data structure)
constant first.

Lines~\ref{alg:nn_2} to~\ref{alg:nn_3} describes how a $node$ is added to
$nodes$. First, we compute the index of $nodes$ array ($iy$) where the $node$
should be stored. Then, the $node$ is added to the list of nodes at that $iy$
index.

When looking for the nearest neighbor of the \(node\) in the indexing structure
(lines~\ref{alg:nn_4} to~\ref{alg:nn_5}), we compute $iy$ index again. Then, we
search the list of nodes stored in the array $nodes$ on index $iy$
($nodes[iy]$). Finally, we repeatedly widen the interval of indexes to be
investigated while the minimum cost is higher than half of the interval width
times $IYSTEP$ and search the lists of nodes stored in the array on indexes
corresponding to the widened interval.

We use Euclidean distance as the cost function in the nearest neighbor search
in contrast to Reeds and Shepp path length as the cost function for building
RRT*.  This approach speeds up the process but does not influence the final
path cost as discussed in Section~\ref{sec:ce}.

\subsection{Path Optimization}
\label{sec:po}

\begin{algorithm}[!htbp]
        \caption{Path optimization}
        \label{alg:opt_path}
        \begin{algorithmic}[1]
        \State Input:\begin{itemize}
                \item path connecting initial and goal pose
        \end{itemize}
        \State Output:\begin{itemize}
                \item lower cost path connecting initial and goal pose
        \end{itemize}
                \Procedure{OptPath}{}
                \State $tips\gets$ cusp nodes\Comment{array}\label{alg:op_5}
                \State $pq\gets \emptyset$\Comment{priority queue}
                \State $pq\gets pq \cup tips[0]$\label{alg:op_6}
                \While{$|pq| \neq 0$}\label{alg:op_7}
                        \State $n_i\gets \Call{Pop}{pq}$
                        \If{$n_i = tips[\Call{Size}{tips} - 1]$}
                                \State \textbf{break}
                        \EndIf
                        \For{all $j > i$}\label{alg:op_1}
                                \State $n_j\gets tips[j]$
                                \State $P_{n_i,n_j}\gets \Call{Steer}{n_i,n_j}$
                                \State $c\gets n_i.ccost+\Call{Cost}{n_i, n_j}$
                                \If{$\Call{Collide}{n_i, n_j}$}
                                        \State \textbf{continue}
                                \EndIf
                                \If{$c < n_j.ccost$}
                                        \State $n_j.ccost\gets c$
                                        \State $n_j.parent\gets i$
                                        \If{$n_j.visited = False$}
                                                \State $n_j.visited\gets True$
                                                \State $pq\gets pq \cup n_j$
                                        \EndIf
                                \EndIf
                        \EndFor\label{alg:op_2}
                \EndWhile\label{alg:op_8}
                \State $opath\gets \emptyset$\Comment{new optimized path}
                \label{alg:op_3}
                \State $i\gets \Call{Size}{tips} - 1$
                \While{$i > 0$}
                       \State $opath\gets opath \cup tips[i]$
                       \State $i\gets tips[i].parent$
                \EndWhile
                \State $opath\gets opath \cup tips[0]$\label{alg:op_4}
                \If{better cost of $opath$}
                        \State \textbf{return} True
                \EndIf
                \State \textbf{return} False
                \EndProcedure
        \end{algorithmic}
\end{algorithm}

\noindent The path optimization procedure is run when the goal is found. Even
that RRT* is asymptotically optimal, it converges to the optimal solution very
slowly. When the \(p_{goal}\) is reached for the first time a final path
\(P_F\) is probably far from optimum in the sense of cost (we use Reeds and
Shepp path length as cost). The purpose of path optimization procedure is to
decrease the final path cost.

A final path \(P_F\) consists of topologically ordered nodes (see definition of
a path in Section~\ref{sec:problem}). We select {\em tip} nodes from the final
path that are also topologically ordered. In our case, {\em tip} nodes are cusp
nodes (nodes where the direction of movement changes) along with \(p_{init}\)
and \(p_{goal}\).

In Algorithm~\ref{alg:opt_path} we initialize {\em tip} nodes and priority
queue in lines~\ref{alg:op_5} to~\ref{alg:op_6}. In lines~\ref{alg:op_7}
to~\ref{alg:op_8}, we use Dijkstra algorithm to find the shortest path from the
first {\em tip} node (\(p_{init}\)) to the last one (\(p_{goal}\)). From the
priority queue, we pop the node \(n_i\) (where \(i\) is the index of \(n_i\)
node in \(tips\) array) with the lowest cumulative cost. Then, we call
\textsc{Steer(\(n_i, n_j\))} procedure from \(n_i\) to all \(n_j\) for \(j>i\)
(see lines~\ref{alg:op_1} to~\ref{alg:op_2}) that returns path \(P_{n_i,
n_j}\). If \(P_{n_i, n_j}\) is collision free and cumulative cost of \(n_j\) is
smaller when reached via \(P_{n_i,n_j}\) then parent and cumulative cost of
\(n_j\) are updated, and \(n_j\) is pushed to the priority queue if not visited
already. The process repeats until the priority queue is empty or
\(n_i=p_{goal}\).

Optimized path found by Dijkstra is retrieved in lines~\ref{alg:op_3}
to~\ref{alg:op_4}. If the cumulative cost of \(p_{goal}\) is better,
\textsc{OptPath} procedure returns \(True\) and \(False\) otherwise.

\section{\uppercase{Computational Experiments and Evaluation}}
\label{sec:ce}

\noindent We present the results of computational experiments for parallel
parking scenario with no obstacle in Section~\ref{sec:ce:no}, and the results
of computational experiments for parallel parking scenarios with circle
obstacle in Section~\ref{sec:ce:wo}.

We are interested in the {\em nearest neighbor} search and {\em path
optimization} procedures. Specifically, we are interested in how does the cost
function, used in the nearest neighbor search, influences algorithm computation
time. We experimented with the following implementations of the nearest
neighbor search:

\begin{itemize}
        \item Nearest neighbor search with the cost based on Reeds and Shepp
        path length.
        \item Nearest neighbor search with the cost based on Reeds and Shepp
        path length but with the heading of nodes temporarily set to the same
        value.
        \item Nearest neighbor search with the cost based on Euclidean
        distance.
\end{itemize}

Also, we would like to know if the path optimization procedure influences the
cost of the final path. We tested the following path optimization
possibilities:

\begin{itemize}
	\item No path optimization.
	\item Path optimization from~\cite{Islam2012}.
	\item Path optimization described in Algorithm~\ref{alg:opt_path}.
\end{itemize}

The car we use for experiments is \SI{1.625}{m} wide and \SI{3.760}{m} long.
The minimum turning radius of the car is \SI{10.820}{m} and wheelbase is
\SI{2.450}{m}.

We run computational experiments on a~single core of Intel(R) Core(TM) i7-5600U
CPU @ \SI{2.60}{GHz} with MemTotal:~\SI{16322516}{kB}.

\subsection{Scenario with No Obstacle}
\label{sec:ce:no}

\begin{figure}[!htbp]
        \vspace{-0.2cm}
        \centering{
         \includegraphics[width=\linewidth]{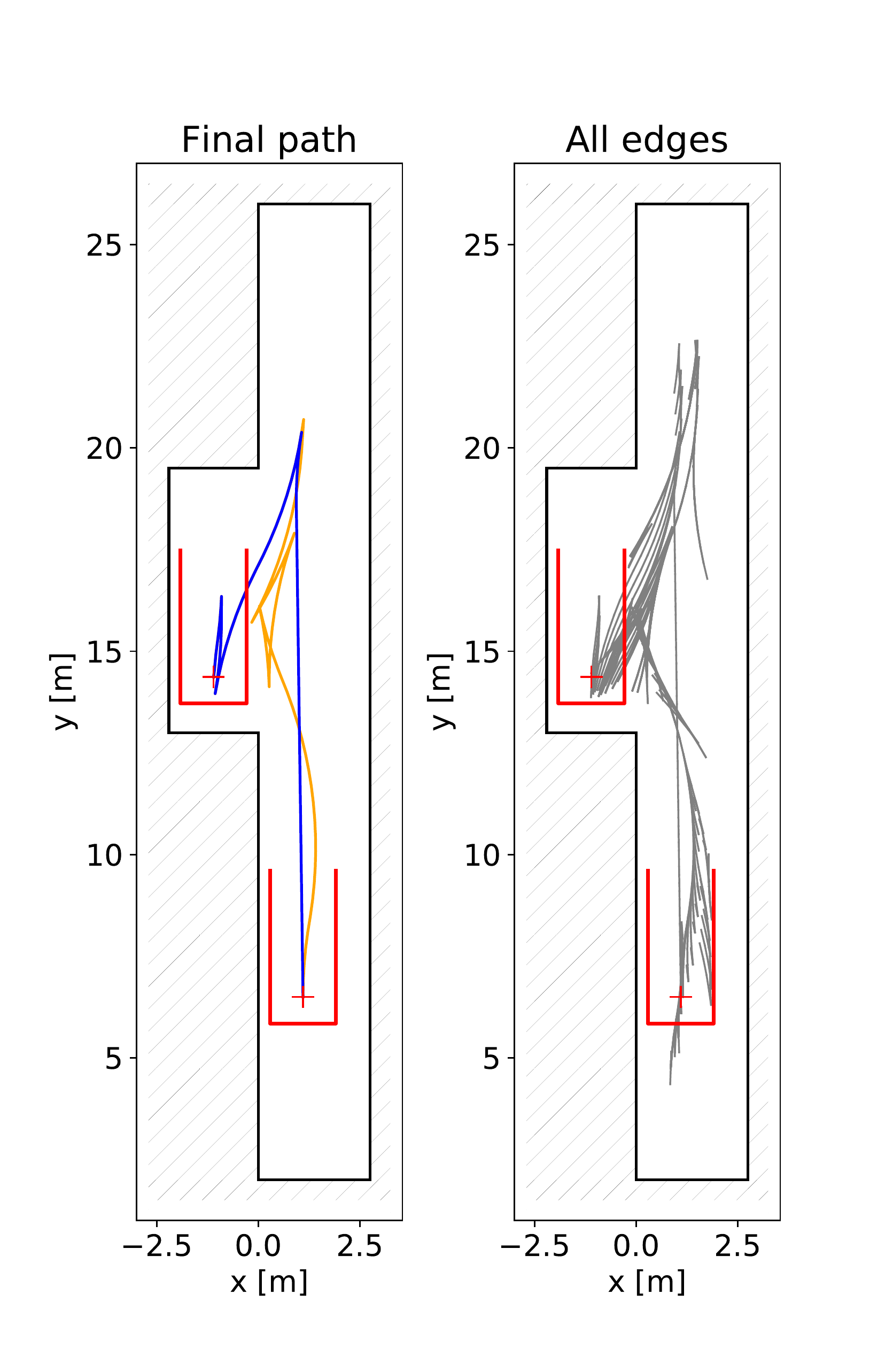}}
        \caption{Parallel parking scenario with no obstacle. On the left, there
        is a final path before optimization (orange) and optimized final path
        (blue). On the right, there is a complete tree of all paths (gray).}

        \label{fig:lpar-no_rrts}
        \vspace{-0.1cm}
\end{figure}

\noindent We tested RRT*-based algorithm on parallel parking scenario shown in
Figure~\ref{fig:lpar-no_rrts}. The parking lot is \SI{2.2}{m} wide and
\SI{6.5}{m} long~\cite{CSN736056}. The width of the street is \SI{2.75}{m}.

We let the Algorithm~\ref{alg:rrtstar} to run for up to 10 seconds. When the
RRT*-based algorithm finds the goal, the \textsc{OptPath} procedure optimizes
the final path. We repeated the experiment \SI{10000}{} times for this
scenario.

\subsubsection{Nearest Neighbor Search}

\begin{figure}[!htbp]
        \vspace{-0.2cm}
        \centering{
         \includegraphics[width=\linewidth]{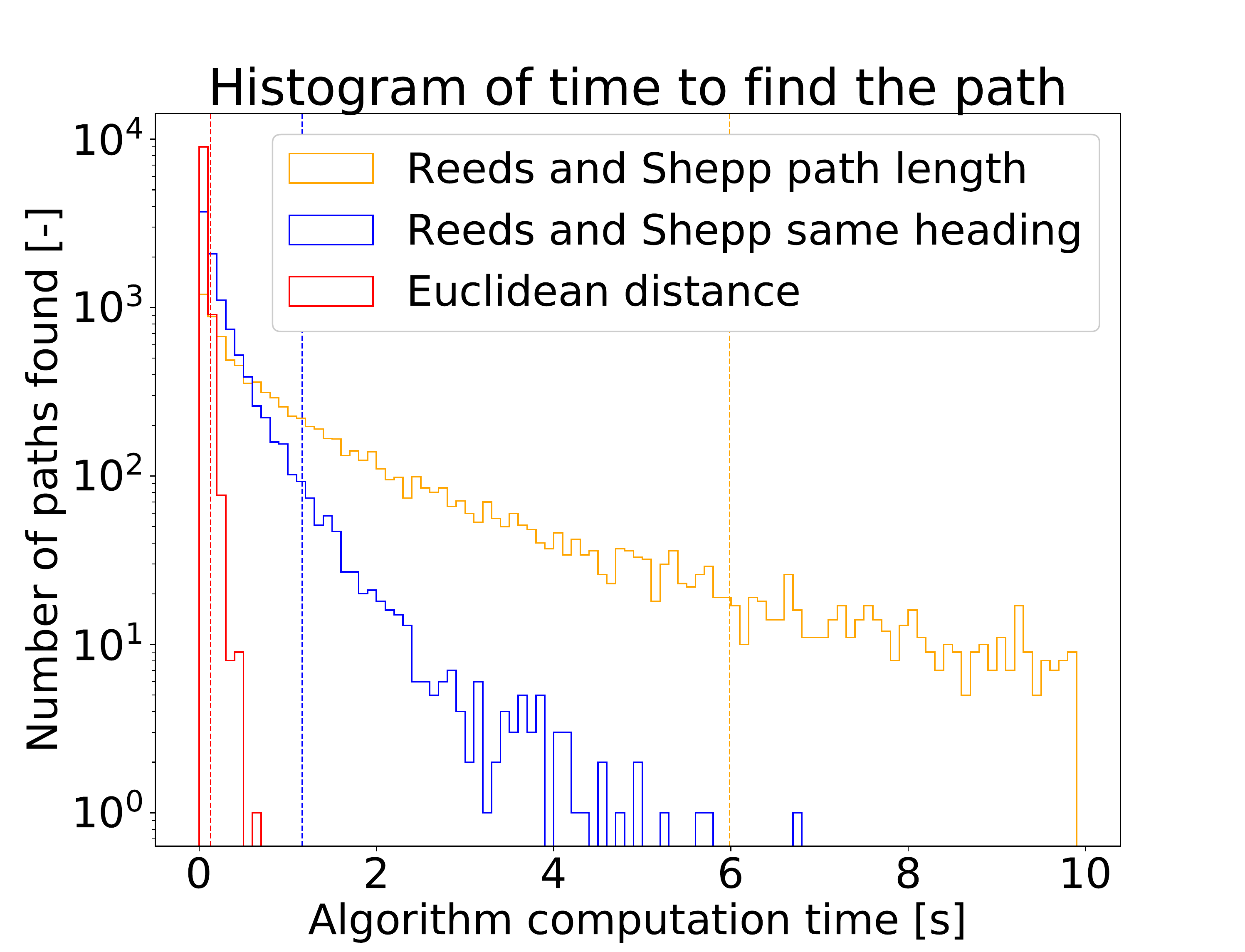}}
        \caption{Histogram of time to find the path. Vertical dashed lines
        represent \SI{95}{\%} percentile (red is 0.13, blue is 1.16, orange is
        5.98).}

        \label{fig:noth}
        \vspace{-0.1cm}
\end{figure}

\noindent We compare the computation times when the algorithm found the final
path for different cost functions used in the nearest neighbor search
implementations. We can see the results in the histogram with the logarithmic
scale in Figure~\ref{fig:noth}.

For the nearest neighbor search implementation with the Reeds and Shepp cost
function (the same cost function used for building \(\mathcal{T}\), {\em
orange} in Figure~\ref{fig:noth}), the algorithm did not find the goal in all
runs. On the other hand, for the nearest neighbor search implementation where
we used the Euclidean distance as the cost function ({\em red} in
Figure~\ref{fig:noth}), the goal was found in \SI{100}{\%} of runs. For
comparison purposes, we run the experiment for the nearest neighbor search
implementation with Reeds and Shepp cost function, where the heading of the
nodes was temporarily set to the same value ({\em blue} in
Figure~\ref{fig:noth}).

\subsubsection{Path Optimization}

\begin{figure}[!htbp]
        \vspace{-0.2cm}
        \centering{
         \includegraphics[width=\linewidth]{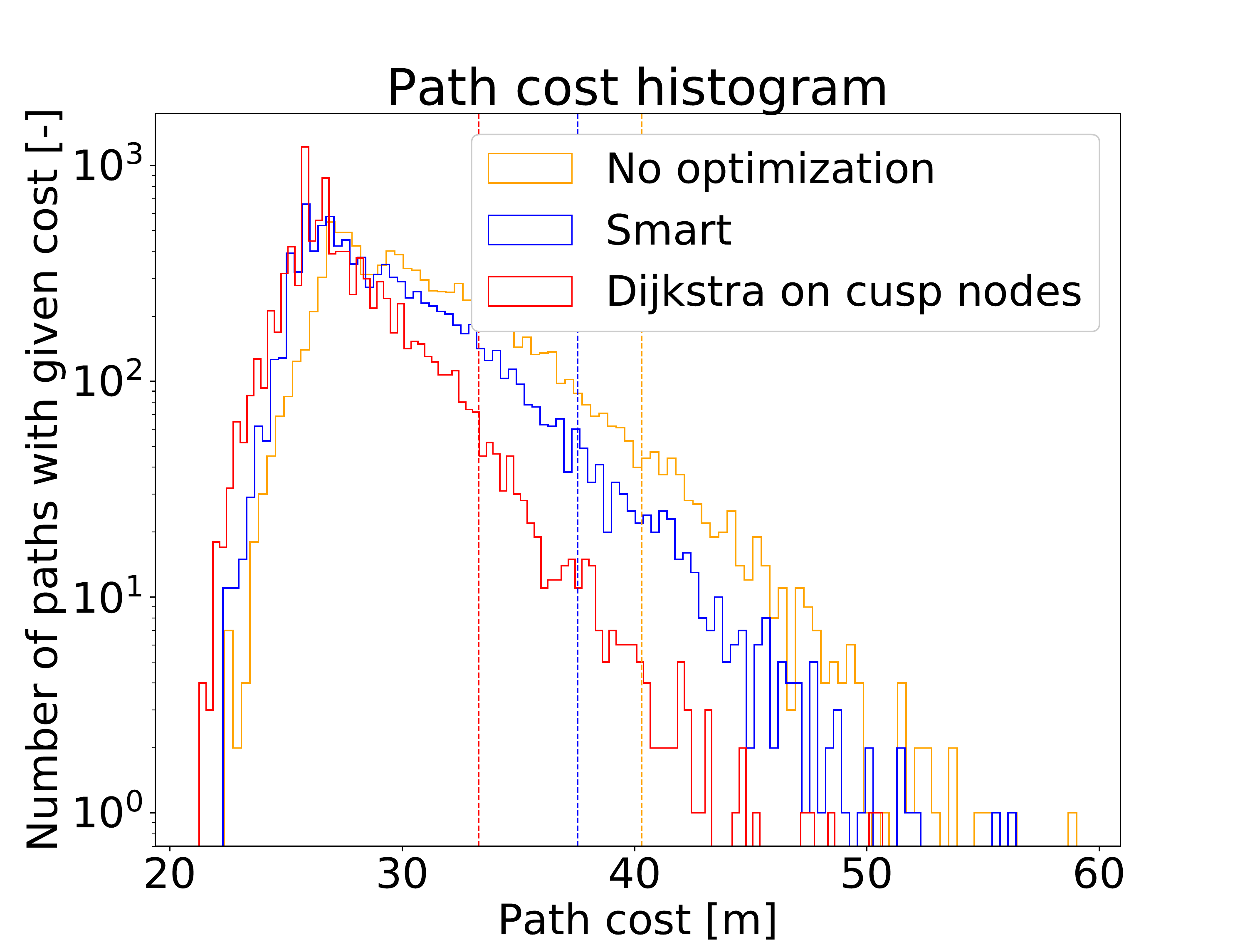}}
        \caption{Path cost histogram. Vertical dashed lines represent
		\SI{95}{\%} percentile (red is 33.29, blue is 37.56, orange is
                40.31).}

        \label{fig:noch}
        \vspace{-0.1cm}
\end{figure}

\noindent We also compared the final path costs for different path optimization
procedures. We can see the results in the histogram with the logarithmic scale
in Figure~\ref{fig:noch}.

We can see the improvement over no path optimization ({\em orange}) when
algorithm from~\cite{Islam2012} is used ({\em blue}). And we can see that the
path optimization from Algorithm~\ref{alg:opt_path} ({\em red}) has the best
results.

\subsection{Scenario with Circle Obstacle}
\label{sec:ce:wo}

\begin{figure}[!htbp]
        \vspace{-0.2cm}
        \centering{
         \includegraphics[width=\linewidth]{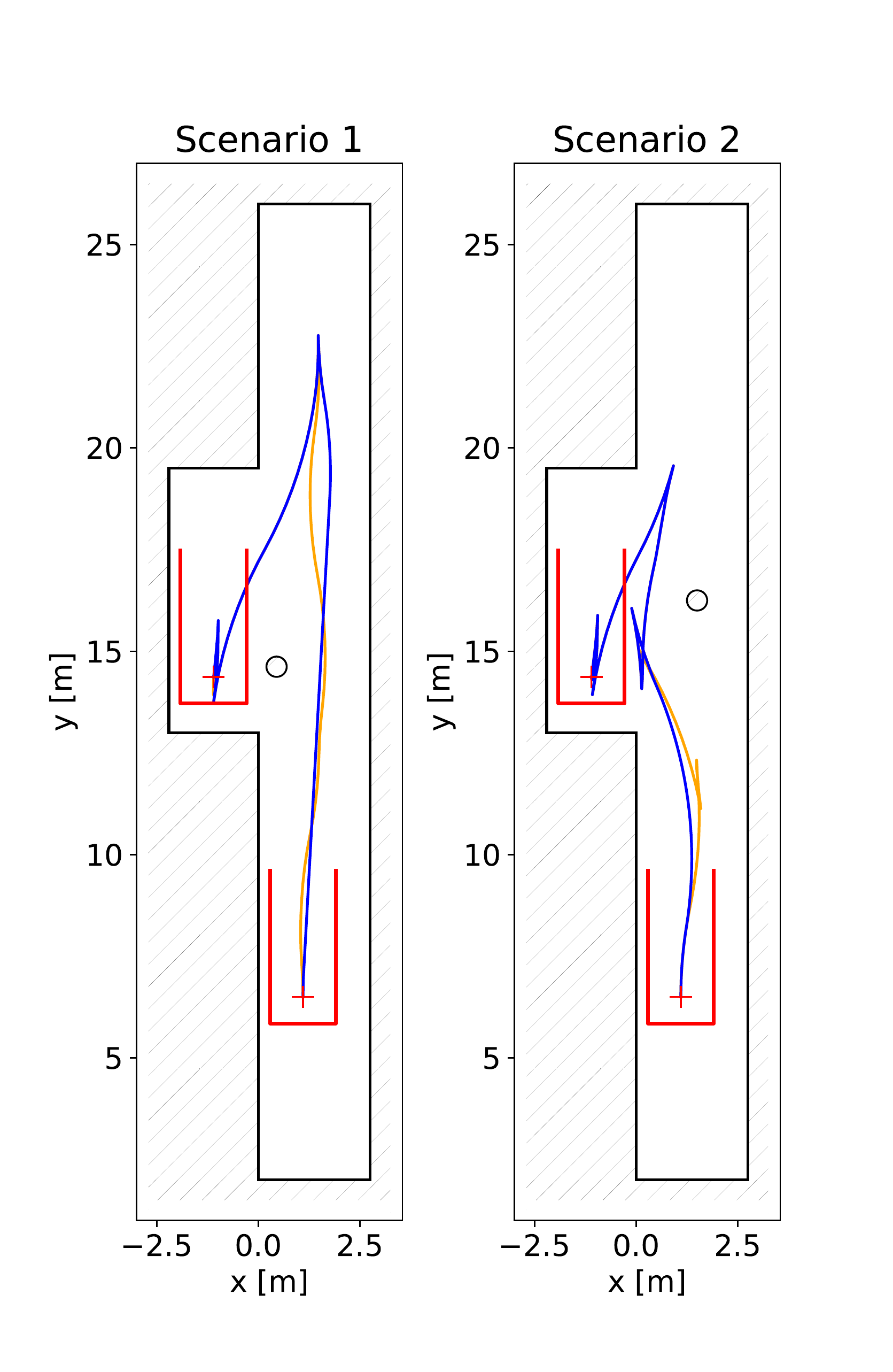}}
        \caption{Parallel parking scenarios with circle obstacle. Scenarios 1
        and 2 differ in the position of circle obstacle. There is the final
        path before optimization (orange) and the optimized final path (blue).}

        \label{fig:lpar-wo_rrts}
        \vspace{-0.1cm}
\end{figure}

\noindent Further, we tested RRT*-based algorithm on parallel parking scenarios
shown in Figure~\ref{fig:lpar-wo_rrts}. The parking lot is \SI{2.2}{m} wide and
\SI{6.5}{m} long~\cite{CSN736056}. The width of the street is \SI{2.75}{m}.
There is a~random circle obstacle with diameter of \SI{0.5}{m} laying on the
street near the parking lot.

We let the Algorithm~\ref{alg:rrtstar} to run for up to 10 seconds. When the
RRT*-based algorithm finds the goal, the \textsc{OptPath} procedure optimizes
the final path. We repeated the experiment \SI{10000}{} times for this
scenario.

The results are similar to the results in Section~\ref{sec:ce:no}. The cost
based on the Euclidean distance speeds up the algorithm computation time
(\SI{95}{\%} percentile), dependent on the obstacle position, to \SI{0.12}{s}
for Scenario 1 and to \SI{0.66}{s} for Scenario 2. The path optimization
procedure decreases the cost of the final path (\SI{95}{\%} percentile) by
\SI{4}{\%} concerning No optimization case for both scenarios.

\section{\uppercase{Conclusion}}
\label{sec:concl}

\noindent We proposed the RRT*-based algorithm for planning parking paths and
experimented with the nearest neighbor search and path optimization procedures.

RRT*-based algorithm without improvements uses the cost function based on the
Reeds and Shepp path length in the nearest neighbor search (as well as for
building the \(\mathcal{T}\) data structure), and no optimization procedure.
Our improvements include the cost function based on Euclidean distance in the
nearest neighbor search and optimization procedure based on the Dijkstra
algorithm.

We have shown that when we use the cost function based on the Reeds and Shepp
path length for building the \(\mathcal{T}\) data structure and the cost
function based on the Euclidean distance in the nearest neighbor search, there
is a significant acceleration in algorithm computation time.

Additionally, we have shown that the path optimization procedure based on the
Dijkstra algorithm for the shortest path search can optimize the final path to
\SI{63}{\%} of the original cost in \SI{95}{\%} of cases, for the parallel
parking scenario without obstacles, which is a better result than the
optimization procedure used in~\cite{Islam2012}. However, for parallel parking
scenario with circle obstacle, the optimized cost is only \SI{96}{\%} of the
original cost in \SI{95}{\%} of cases.

Finally, from the experiments we can see that for parallel parking scenario
with no obstacle, RRT*-based algorithm with improvements tends to significantly
faster computation time as well as to lower final path cost. However, for
parallel parking scenario with circle obstacle, RRT*-based algorithm with
improvements tends to significantly faster computation time but about
\SI{10}{\%} to \SI{20}{\%} worse final path cost then RRT*-based algorithm
without improvements.

In our future work, we are going to experiment with the improvements presented
in this paper. Particularly, the recognition and selection of {\em tip} nodes
seem to be interesting. Also, the bidirectional RRT* algorithms, such as
\cite{Jordan2013} and \cite{Klemm2015}, could lead to significant improvements
in the matter of computational time.

\section*{\uppercase{Acknowledgements}}

\noindent This work was supported by the Technology Agency of the Czech
Republic under the Centre for Applied Cybernetics TE01020197.

\vfill
\bibliographystyle{apalike}
{\small
\bibliography{lib}}

\vfill
\end{document}